\documentclass[sigconf]{acmart}

\usepackage{booktabs} 
\usepackage{amsmath, amsthm, amssymb}
\usepackage{comment}
\usepackage{multirow}

\usepackage{rotating}
\usepackage{tabularx}
\usepackage{caption}
\usepackage[linesnumbered,ruled,vlined]{algorithm2e}

\usepackage{pgfplots}
\usetikzlibrary{patterns}
\usetikzlibrary{matrix}
\usepackage{enumitem}
\usetikzlibrary{fit,positioning}
\usepackage{subfigure}
\usepackage{tikz}
\usepackage{tikz-qtree}
\usepackage{pgfplots}
\usetikzlibrary{patterns}

\copyrightyear{2018}
\acmYear{2018} 
\setcopyright{iw3c2w3}
\acmConference[WWW 2018]{The 2018 Web Conference}{April 23--27, 2018}{Lyon, France}
\acmBooktitle{WWW 2018: The 2018 Web Conference, April 23--27, 2018, Lyon, France}
\acmPrice{}
\acmDOI{10.1145/3178876.3186033}
\acmISBN{978-1-4503-5639-8/18/04}

\begin{document}
\title{Leveraging Crowdsourcing Data For Deep Active Learning\\ An Application: Learning Intents in Alexa}
\renewcommand{\shorttitle}{Leveraging Crowdsourcing Data For Deep Active Learning}
\author{Jie Yang}
\authornote{This work was done while the first author was an intern at Amazon Research.}
\affiliation{
  \institution{Delft University of Technology}
  \city{Delft} 
  \country{Netherlands} 
}
\email{j.yang-3@tudelft.nl}

\author{Thomas Drake}
\affiliation{
  \institution{Amazon Research}
  \city{Seattle, WA} 
  \country{USA} 
}
\email{draket@amazon.com}

\author{Andreas Damianou}
\affiliation{
  \institution{Amazon Research}
  \city{Cambridge} 
  \country{UK} 
}
\email{damianou@amazon.com}

\author{Yoelle Maarek}
\affiliation{
  \institution{Amazon Research}
  \city{Haifa} 
  \country{Israel} 
}
\email{yoelle@yahoo.com}



\begin{abstract}
This paper presents a generic Bayesian framework that enables any deep learning model to actively learn from targeted crowds. Our framework inherits from recent advances in Bayesian deep learning, and extends existing work by considering the targeted crowdsourcing approach, where multiple annotators with unknown expertise contribute an uncontrolled amount (often limited) of annotations. Our framework leverages the low-rank structure in annotations to learn individual annotator expertise, which then helps to infer the true labels from noisy and sparse annotations. It provides a unified Bayesian model to simultaneously infer the true labels and train the deep learning model in order to reach an optimal learning efficacy. Finally, our framework exploits the uncertainty of the deep learning model during prediction as well as the annotators' estimated expertise to minimize the number of required annotations and annotators for optimally training the deep learning model. 

We evaluate the effectiveness of our framework for intent classification 
in Alexa (Amazon's personal assistant), using both synthetic and real-world datasets. Experiments show that our framework can accurately learn annotator expertise, infer true labels, and effectively reduce the amount of annotations in model training as compared to state-of-the-art approaches. We further discuss the potential of our proposed framework in bridging machine learning and crowdsourcing towards improved human-in-the-loop systems.
\end{abstract}

%
%




\maketitle
\section{Introduction}

Deep learning models have achieved remarkable success for automating various tasks, ranging from image recognition \cite{krizhevsky2012imagenet}, speech recognition \cite{graves2013speech}, to natural language processing \cite{sutskever2014sequence}. These models often require a large number of parameters, significantly greater than classic machine learning models, in order to capture complex patterns in the data and thus to achieve superior performance in prediction tasks \cite{zhang2016understanding}.  
Learning these parameters, however, typically requires large amount of labeled data. In fact, researchers have identified strong correlations between the capability of deep learning models, the number of parameters in the model, and the size of the training data \cite{Goodfellow-et-al-2016}. 
Obtaining these labels is a long, laborious, and usually costly process. 
Crowdsourcing provides a convenient means for data annotation at scale. For example, the ImageNet dataset \cite{deng2009imagenet} -- one of the most popular datasets driving the advancement of deep learning techniques in computer vision -- is annotated by 49K workers recruited from Amazon Mechanical Turk\footnote{\url{https://www.mturk.com/}} over 3 years (2007-2010) for  3.2M images. In practice, data annotation for training machine learning models is one of the main applications of crowdsourcing~\cite{law2011human,demartini2015hybrid,raykar2010learning,lease2011quality}.  


As of today, data annotation and model training are generally regarded as isolated processes. Task owners collect annotations from a supposedly unlimited source of annotators who are assumed to be anonymous and disposable, and then train the deep learning model for the application at hand. This assumption, however, does not hold for many tasks that are either subjective or knowledge intensive. An example of such a task, is the one we are considering in this paper: intent classification of users' queries in conversational agents. Such a task is key to the effectiveness of personal assistants such as Amazon Alexa\footnote{\url{https://developer.amazon.com/alexa}} or Google Home\footnote{\url{https://madeby.google.com/home/}}. The true query intent is highly subjective, and is largely dependent on various contextual factors. 
As a result, annotations generated by anonymous, even if trained, workers cannot be fully trusted. 

Crowdsourced data annotation requires certain type of workers. In our conversational agent application, the ideal annotators are the users who issued the queries. This is feasible only if the task allows for a natural, non obtrusive way for users to confirm their query intents, e.g., Alexa requires the user to confirm their intents by responding to the question ``do you want to shop?''. 
This type of crowdsourcing, referred to \emph{targeted crowdsourcing}, has been studied by Ipeirotis et al. \cite{ipeirotis2014quizz}. Unlike the conventional notion of paid crowdsourcing, targeted crowdsourcing features a certain group of annotators of varying expertise. This leads to several challenges including identifying the right annotators and effectively engaging them. In this paper, we investigate \emph{how to best train a deep learning model while minimizing the data annotation effort in targeted crowdsourcing}. This problem is important given the targeted annotators as valuable worker resources. 
In our conversational agent context, minimizing the amount of confirmation questions for users is critical to reduce the negative impact on customer experience. 


We propose here to adopt an \emph{active learning} approach, as it allows the model to choose the data from which it learns  best~\cite{settles2010active}. With active learning, models that are initially trained on a small dataset actively make decisions in order to select the most informative data samples, often based on the model's uncertainty. These data samples are then routed to \emph{an expert} for annotation, and inserted into the training set for model retraining. With active learning, we can expect the model to be effectively trained with the minimum amount of contribution from crowds. Moreover, by performing active learning over time, the model can detect the changes in application environment and adapt accordingly, thereby continuously delivering high-quality prediction. 

Despite the potential, enabling deep learning models to actively learn from targeted crowds is non-trivial for several reasons. First, deep learning models can rarely represent prediction uncertainty -- they usually perform prediction in a deterministic way. 
Second, while targeted crowds can provide better labels than explicitly recruited workers, annotation quality remains a significant issue. For example, annotators' expertise (e.g., users' familiarity with the application) can have a highly influential impact on the annotation quality. Furthermore, users may not even answer the confirmation question asked by conversational agents, or may give random responses when losing interest, which happens frequently in reality. Last but not least, annotations provided by targeted crowds can be highly sparse, making it particularly challenging to model annotator expertise or annotation quality. In contrast to conventional crowdsourcing, where requesters have control over the number of annotations contributed by individual workers, most annotators in targeted crowdsourcing may only contribute a small number of annotations due to the lack of engagement mechanisms.



To address these issues, this paper introduces a generic Bayesian framework that supports effective deep active learning from targeted crowds. Our framework inherits from recent advances in Bayesian deep learning \cite{gal2016dropout,welling2011bayesian}, and leverages dropout as a practical way for representing model uncertainty in deep learning. 
To resolve the annotation noise and sparsity issues, our framework exploits the low-rank structure in annotations and learns a low-dimensional representation of individual annotator expertise, which is then used to learn annotation reliability, so as to reduce annotation noise. 
Annotation reliability is further learned in a way conditioned on specific data samples, so that samples that are intrinsically more ambiguous can be identified. 
Using a Bayesian approach, our framework simultaneously infers the true label from noisy and sparse crowd annotations and trains the deep learning model to reach an optimal learning efficacy. 
This approach brings an additional benefit: the annotator selection process and the network training process influence each other, allowing the active learning to be tailored to the hidden representation and objective function of the neural network. This comes in contrast to schemes used in practice that employ one model with well-calibrated uncertainty (e.g., Gaussian processes \cite{krause2008near}) for active learning and a different model (e.g., a deep neural network) for prediction. In active learning settings, our framework exploits the uncertainty of the deep learning model in prediction as well as the learned annotator expertise in order to respectively minimize the number of required annotations and the number of annotators for optimally training the deep learning model. 

The key contributions  of our work include:
\begin{itemize}[leftmargin=*]
\item Introducing the notion of deep active learning in targeted crowdsourcing settings.
\item Proposing a method for learning annotator expertise and inferring true labels from noisy and sparse crowd annotations, which takes advantage of the low-rank structure of the annotations.
\end{itemize}
\begin{itemize}[noitemsep,nolistsep,leftmargin=*]
\item Defining a generic Bayesian framework that learns annotator expertise, infers true labels, and trains the deep learning model simultaneously. This framework further reduces annotation efforts to enable deep learning models to actively learn from crowds.
\item Validating our approach and framework via extensive experiments on both synthetic and real-world datasets. In particular, we demonstrate the effectiveness of our framework in Amazon's Alexa, one of today's major conversational agents. 
\end{itemize}

\smallskip
To the best of our knowledge, this work is the first to study deep active learning from sparse and noisy crowd annotations. Our proposed framework is a generic one applicable to any deep learning model and in a variety of domains, e.g., natural language processing and computer vision.  



\section{Related Work}
In this section, we first discuss relevant work from the emerging field of human-in-the-loop paradigm, then we review existing work methodologically related to our proposed framework in Bayesian deep active learning and learning from crowds.

\subsection{Human-in-the-Loop Systems}
Human-in-the-loop systems, or human-machine hybrid systems, are aimed at exploiting the complementarity between the intelligence of humans and the scalability of machines to solve complex tasks at scale \cite{demartini2015hybrid}. A number of human-in-the-loop systems have been proposed up to date. These include an early example of the ESP game \cite{von2008designing}, and systems that demonstrate the amplified power of human intelligence when coupled with machines in solving complex tasks for automated systems (e.g., Recaptcha for OCR applications~\cite{von2008recaptcha}). More recent human-in-the-loop systems have been proposed to solve data-related problems in a variety of domains. For example, CrowdDB by Franklin et al. \cite{franklin2011crowddb} for Databases,  CrowdSearcher by Bozzon et al. \cite{bozzon2012answering} for Information Retrieval, and Zencrowd by Demartini et al. \cite{demartini2012zencrowd} for Semantic Web.

An important application area of human-in-the-loop systems is machine learning, by engaging crowds to annotate data for training supervised machine learning models. 
Examples include the ImageNet dataset for computer vision \cite{deng2009imagenet} and many crowd annotated datasets for various natural language processing tasks, such as datasets for sentiment and opinion mining \cite{mellebeek2010opinion} and  question answering \cite{heilman2010rating}, to name a few. 
An often neglected aspect of crowdsourced data annotation for machine learning is that data annotation and model training are generally regarded as isolated processes. This does not work well for subjective or knowledge-intensive tasks, for which workers are regarded as a valuable resource, with only specific workers able to provide high-quality annotations. A notion developed for this particular type of crowdsourcing is called \emph{targeted crowdsourcing} by Ipeirotis and Gabrilovich \cite{ipeirotis2014quizz}, which emphasizes the demand for worker expertise. A similar concept has been proposed as \emph{nichesourcing} by De Boer et al. \cite{de2012nichesourcing}. How to best train machine learning models while minimizing workers' efforts in targeted crowdsourcing remains an open question. The problem becomes even more challenging for deep neural network based machine learning models (i.e., deep learning models) that are prevalent in many domains, due to the strong influence of the size of training data on model performance. 

\subsection{Bayesian Deep Active Learning}
To enable deep learning models to actively learn from crowds, we base our approach on Bayesian deep active learning \cite{gal2017deep}, which unifies deep learning with active learning using Bayesian methods. In the following, we briefly review related work that converges to the current notion of Bayesian deep active learning. 

First, we consider the same model for driving the active labeling and the predictive learning task. In such a setting, the model selects unlabeled data samples which can provide the strongest training signal; these samples are labeled and used to supervise model training. The potential benefit of a data sample is generally measured by the model's uncertainty in making predictions for that sample, i.e., the so-called \emph{uncertainty sampling} \cite{lewis1994sequential,cohn1996active}. Other criteria also exist, e.g., how well a data sample will reduce the estimate of the expected error \cite{roy2001toward}, which attempts to select data samples that directly optimize prediction performance. Such a criterion, however, is less practical than model uncertainty as it is generally difficult to have an analytical expression for the expected prediction error. As a remark, the traditional notion of active learning assumes a single omniscient oracle that can provide genuine labels for any data samples, without any constraint on the amount of provided annotations. 

Unlike probabilistic models (e.g., Bayesian networks), deep learning models 
 only make deterministic predictions, making it rather challenging to represent model uncertainty, which is essential for active learning. A popular workaround has been to employ an active learning model (e.g., a Gaussian process) which is separate from the neural network classifier \cite{krause2008near}. A more consistent solution is to leverage recent developments in Bayesian deep learning. 
The Bayesian deep learning methods of interest can be generally categorized into two classes. The first class is based on stochastic gradient descent (SGD). 
Welling et al. \cite{welling2011bayesian,ahn2012bayesian} show that by adding the right amount of noise to the standard SGD the parameter will converge to samples from the true posterior distribution. The other class of methods is based on dropout, which is a technique originally proposed to prevent over-fitting in training deep learning models \cite{srivastava2014dropout}. It is proved by Gal et al. \cite{gal2016dropout} that when preserving dropout during prediction the same way as in model training, the predictions are equivalent to sampling from the approximate true posterior distribution of the parameters, thus turning a deterministic predictive function into a stochastic (uncertain) one. 

A recent paper \cite{gal2017deep}, perhaps the most closely related work to ours, proposes a method for Bayesian deep active learning based on dropout. It follows the conventional assumption of active learning: a single expert is able to provide high-quality annotations on demand. This is unrealistic in practice, especially for deep learning models which would require large amounts of annotated data. To the best of our knowledge, we are the first to investigate a principled approach to enable deep learning models to actively learn from crowds.

\subsection{Learning from Crowds}
While not focusing on deep learning models, there is a line of research \cite{raykar2010learning,yan2011active,tian2012learning,fang2012self,zhou2012learning,zhong2015active,laws2011active} that has investigated methods for enabling machine learning models to learn from crowds. 
The key problems here are two-fold: 1) infer the true label, and 2) train the model. The former problem often requires estimation of the reliability of annotations, which is further related to the expertise of annotators \cite{bozzon2014modeling} and the difficulty or clarity of tasks \cite{yang2016modeling,gadiraju2017clarity,yang2014asking}. 

In an early work, Dawid and Skene \cite{dawid1979maximum} first study the problem of inferring true labels from multiple noisy labels and introduce an Expectation Maximization (EM) algorithm to model worker skills. Sheng et al. \cite{sheng2008get} shows that repeated-labeling by multiple annotators can significantly improve the quality of the labels using a simple majority voting label aggregation scheme. Whitehill et al. \cite{whitehill2009whose} generalize the work from Dawid and Skene by taking into account both worker expertise and task difficulty for label inference, and show better performance than majority voting. 

Raykar et al. \cite{raykar2010learning} first introduce the problem of learning-from-crowds to improve machine learning models. A Bayesian method is proposed to model the true label as a joint function that considers both crowd annotations and the output of a logistic regression classifier. The parameters of annotator expertise and the classifier are then learned by maximizing the likelihood of the observed data and crowd annotations.  Their method, however, does not consider task properties (e.g., task type) as influential factors in annotation reliability. Yan et al. \cite{yan2010modeling} extend the problem by considering the influence of task properties on annotation reliability. They formulate annotation reliability as a logistic function parameterized by both worker and task representations. In a slightly different scenario, Tian et al. \cite{tian2012learning} model the problem of ``schools of thought'', where multiple correct labels can exist for a single data sample. 

A closely related line of research incorporates active learning into learning-from-crowds, to reduce the cost in annotation. Yan et al. \cite{yan2011active} extend their work in \cite{yan2010modeling} to pick most uncertain data samples and most reliable workers for active learning. Fang et al.~\cite{fang2012self} then consider the case when annotators are able to learn from one another to improve their annotation reliability. In a more recent work, Zhong et al. \cite{zhong2015active} further model the scenario when workers can explicitly express their annotation confidence by allowing them to choose an unsure option. 

All these works, however, do not consider deep learning models as the target model to improve. Furthermore, none of them considers the targeted crowdsourcing scenarios where the annotations contributed by individual annotators are both noisy and sparse. 



\section{The DALC Framework}
This section introduces our proposed framework that we refer to as Deep Active Learning from targeted Crowds (DALC). We first formalize the problem, then introduce our method for 1) formulating any deep learning (DL) model in a Bayesian framework, and 2) learning from targeted crowds. We then describe the overall Bayesian framework that seamlessly unifies these two models and actively learns from targeted crowds to reduce the amount of annotations for  training DL models.

\smallskip
\noindent\textbf{Problem Formalization.}
Throughout this paper we use boldface lowercase letters to denote vectors and boldface uppercase letters to denote matrices. For an arbitrary matrix $\mathbf{M}$, we use $\mathbf{M_{ij}}$ to denote the entry at the $i$-th row and $j$-th column, and use $\mathbf{M_{i:}}$ to denote the vector at the $i$-th row. We use capital letters (e.g., $\mathcal{P}$) in calligraphic math font to denote sets. 
Let $\mathcal{X} = \{\mathbf{x}_1, \mathbf{x}_2, \ldots, \mathbf{x}_m\}$ ($\mathbf{x}_i\in \mathbb{R}^k$) denote $m$ data samples labeled by $n$ annotators $\mathcal{U} = \{u_1, u_2, \ldots, u_n\}$. The labels form a sparse matrix $\mathbf{L}\in \mathbb{R}^{m\times n}$ where $\mathbf{L}_{ij}$ is the label for sample $\mathbf{x}_i$ contributed by annotator $u_j$. The true labels -- which are unknown -- are denoted as $\mathcal{Y} = \{y_1, y_2, \ldots, y_m\}$ for the $m$ data samples. Given the observed data $\mathcal{X}$ and annotations $\mathbf{L}$ contributed by $\mathcal{U}$, our goals is to infer the true labels $\mathcal{Y}$ and train a deep learning model, whose parameters $\mathcal{W}$, i.e., the weight matrix and bias in each layer of the DL model, are to be learned.

\subsection{Bayesian Deep Learning}
DALC adopts the Bayesian approach to deep learning recently developed by Gal and Ghahramani \cite{gal2016dropout}. Specifically, we consider a generic DL model with parameters $\mathcal{W}$ as a likelihood function:
\begin{equation}
p(y_i|\mathbf{x}_i, \mathcal{W}) = \text{softmax}(f^{\mathcal{W}}(x_i))
\end{equation}
with $f^{\mathcal{W}}(x_i)$ modeling the  output of the network layers preceding the softmax layer. To formulate the DL model in a Bayesian framework, we first define a prior over the parameters $\mathcal{W}$:
\begin{equation}
\small
\mathcal{W} \sim p(\mathcal{W}|K)
\end{equation} 
\noindent e.g., a standard Gaussian prior parameterized by $K$ (the co-variance matrix). With this assumption, model training will result in a posterior distribution over the parameters, i.e., $p(\mathcal{W}| \mathcal{D}_{train})$, instead of point estimates, i.e., fixed values  for the parameters. Note that here we assume that the true labels are given in the training data $\mathcal{D}_{train} = (\mathcal{X}, \mathcal{Y})$. We explain in Section~\ref{sec:opt} how to infer the true labels from the observed data samples and noisy annotations. 

\smallskip
\noindent\textbf{Training \& Prediction with Bayesian DL.} In the following, we describe how to train the Bayesian DL model and make predictions with the trained model. These will be the building blocks in training the overall DALC framework, as we will show in Section~\ref{sec:opt}.

Training the Bayesian DL model is exactly similar to training a  normal DL model using the back-propagation method with dropout enabled. The prediction given an arbitrary input $\mathbf{x}$ can be described as a likelihood function: 
\begin{equation}
\small
p(y|\mathbf{x}, \mathcal{D}_{train}) = \int p(y|\mathbf{x}, \mathcal{W}) p(\mathcal{W}| \mathcal{D}_{train}) d\mathcal{W}
\label{equ:pred}
\end{equation}
\noindent which provides a more robust prediction than non-Bayesian method as it considers the uncertainty of the learned parameters. 

The problem of inferring the exact posterior distribution for the parameters, $p(\mathcal{W}|\mathcal{D}_{train})$, is almost always infeasible in Bayesian methods. Gal and Ghahramani \cite{gal2016dropout} propose Monte Carlo (MC) dropout, which is a simple yet effective method for performing approximate variational inference. 
MC dropout is based on dropout \cite{hinton2012improving,srivastava2014dropout}, which is a widely used method in training DL models to prevent over-fitting. It is generally used during the model training process by randomly dropping hidden units of the network in each iteration. Gal and Ghahramani \cite{gal2016dropout} prove that by simply performing dropout during the forward pass in making predictions, the output is equivalent to the prediction when the parameters are sampled from a variational distribution of the true posterior. To give the intuition, the reason for the above is that test dropout gives us predictions from different versions of the network, which is equivalent to taking samples from a stochastic version of the network. Uncertainty can then be estimated based on the samples similarly to the Query by Committee \cite{seung1992query} principle, which looks at the degree of disagreement.

Formally, MC dropout is equivalent to sampling from a variational distribution $q(\mathcal{W})$ that minimizes the Kullback-Leibler (KL) divergence to the true posterior $p(\mathcal{W}| \mathcal{D}_{train})$. Given this, we can then perform a Monte Carlo integration to approximate Equation~\ref{equ:pred}:
\begin{equation}
\small
\begin{split}
p(y|\mathbf{x}, \mathcal{D}_{train}) & \approx \int p(y|\mathbf{x}, \mathcal{W}) q(\mathcal{W}) d\mathcal{W} \\
& \approx \frac{1}{T} \sum_{t=1}^T p(y|\mathbf{x}, \widehat{\mathcal{W}})
\end{split}
\label{equ:pred_vi}
\end{equation}
\noindent where $\widehat{\mathcal{W}}$ is sampled $T$ times from $q(\mathcal{W})$, i.e., $\widehat{\mathcal{W}} \sim q(\mathcal{W})$. To summarize, MC dropout provides a practical way to approximately sample from the true posterior without explicitly calculating the intractable true posterior. 

\subsection{Learning from Targeted Crowds}
\label{sec:lftc}
The learning-from-targeted-crowds (LFTC) model formulates the relationship among the following objects: the data sample $\mathbf{x}_i$, the true label $y_i$, the annotator $u_j$, and the noisy annotations $\mathbf{L}_{ij}$. We assume the label $\mathbf{L}_{ij}$ contributed by annotator $u_j$ to the data sample $x_i$ is influenced by all of the following three factors: 
\begin{enumerate}[noitemsep,nolistsep,leftmargin=*]
\item the true label $y_i$ -- approximating the true label with noisy annotations is one of our main goals;
\item the data sample $\mathbf{x}_i$ -- it's a realistic assumption that for some data samples (e.g., more ambiguous) the annotation is more noisy; 
\item the annotator $u_j$ -- the label is also dependent on annotator properties, e.g., expertise.
\end{enumerate}

To formalize the relationships described above, we first  represent each annotator $u_j$ as a low-dimensional embedding vector $\mathbf{u}_j\in \mathbb{R}^d$ where $d\ll \min(m,n)$. Each dimension of $\mathbf{u}_j$ represents a latent topic -- which is to be learned from the data -- and each element in $\mathbf{u}_j$ can be viewed as user $u_j$'s expertise in the corresponding topic. 
We then use a Bernoulli likelihood function to model the reliability of an annotation $\mathbf{L}_{ij}$ w.r.t. the true label $y_i$, parameterized by both the data sample $\mathbf{x}_i$ and the annotator $\mathbf{u}_j$:
\begin{equation}
\small
\begin{split}
p(\mathbf{L}_{ij} | \mathbf{x}_i, \mathbf{u}_j, y_i) & = 
(1-\eta_t(\mathbf{x}_i, \mathbf{u}_j))^{|\mathbf{L}_{ij} - y_i|}\eta_t(\mathbf{x}_i, \mathbf{u}_j)^{1-|\mathbf{L}_{ij} - y_i|}
\end{split}
\label{equ:bernoulli}
\end{equation}
that is, the probability of the annotation being correct is  $\eta(\mathbf{x}_i, \mathbf{u}_j)$, which is defined as follows:
\begin{equation}
\small
\begin{split}
\eta(\mathbf{x}_i, \mathbf{u}_j) & = (1+\exp(-\mathbf{u}_j^\intercal \mathbf{F}\mathbf{x}_i))^{-1}
\end{split}
\label{equ:eta}
\end{equation}
where $\mathbf{F}\in \mathbb{R}^{d\times k}$ is a parameter matrix to be learned. $\mathbf{F}$ is as a linear operator that transforms a data sample $\mathbf{x}_i$ of an arbitrary dimension $k$ to an embedding of a low-dimension $d$, i.e., $\mathbf{F}\mathbf{x}_i \in \mathbb{R}^d$, which is then combined with the annotator embedding $\mathbf{u}_j$ via an inner product, to ultimately represent the reliability of the annotation. Intuitively, $\mathbf{F}\mathbf{x}_i$ can be interpreted as the representation of the data sample $\mathbf{x}_i$ by latent topics: each element $\mathbf{F}\mathbf{x}_i$ represents the extent to which the data sample belongs to one of $k$ latent topics. Considering the annotator's expertise over different latent topics represented by $\mathbf{u}_j$, the product between $\mathbf{F}\mathbf{x}_i$ and $\mathbf{u}_j$ can therefore be interpreted as the reliability of $\mathbf{u}_j$'s annotation to $\mathbf{x}_i$. We use sigmoid function in Equation~\ref{equ:eta} to bound reliability between 0 and 1, with 0 standing for completely unreliable and 1 for fully reliable. 

With Equation~\ref{equ:eta}, we could obtain a reliability score $\eta(\mathbf{x}_i, \mathbf{u}_j)$ for each individual annotation $\mathbf{L}_{ij}$. Therefore, the sparse annotation matrix $\mathbf{L}_{ij}$ will result in a sparse matrix whose entries are the corresponding reliability scores (learning these scores is formally described in Section~\ref{sec:opt}). Given the sparsity of the annotation matrix, the low-dimensional assumption of the representations of the annotators $\mathbf{u}_j$ and the data sample $\mathbf{F}\mathbf{x}_i$ is not only critical to effectively learn annotation reliability, but also realistic -- the number of latent topics are often much smaller than the number of annotators and data samples.

Following the Bayesian approach, we assume that annotator embeddings are generated from a prior Gaussian probability, namely:
\begin{equation}
\small
\mathbf{u}_j \sim \mathcal{N}(0, \sigma^2\mathbf{I})
\end{equation}
where $\sigma^2$ is variance and $\mathbf{I}$ is the identity matrix. This prior regularizes the latent topics, which helps to improve the robustness of the model.

\smallskip
\noindent\textbf{Remarks.} Without loss of generality, the above formulation considers the binary classification case. It can be easily extended to a multi-class classification setting, by modeling  $\mathbf{L}_{ij}$ (and $y_i$) as a vector whose $r$-th entry takes value 1 if the annotation (and true label) is class $r$, and 0 otherwise. We use the following Bernoulli probability to model the reliability of annotations:
\begin{equation}
\small
p(\mathbf{L}_{ij}|\mathbf{x}_i, \mathbf{u}_j, y_i) = (1-\eta(\mathbf{x}_i, \mathbf{u}_j))^{\frac{1}{2}\|\mathbf{L}_{ij} - y_i\|_2^2}\eta(\mathbf{x}_i, \mathbf{u}_j)^{1-\frac{1}{2}\|\mathbf{L}_{ij} - y_i\|_2^2}
\end{equation}

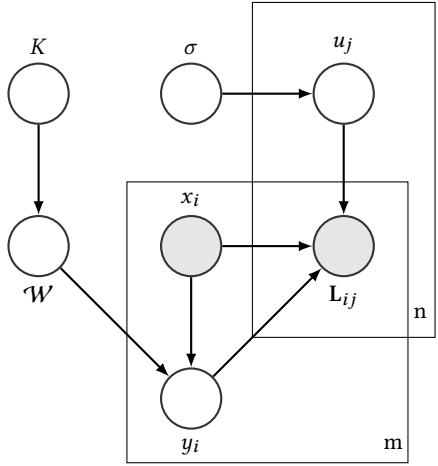
\begin{figure}
\centering
\begin{tikzpicture}
\tikzstyle{main}=[circle, minimum size = 8mm, thick, draw =black!80, node distance = 12mm]
\tikzstyle{connect}=[-latex, thick]
\tikzstyle{box}=[rectangle, draw=black!100]
  \node[main, fill = white!100] (p) [label=above:$K$] { };
  \node[main] (W) [below=of p,label=below:$\mathcal{W}$] { };
  \node[main, fill = black!10] (x) [right=of W,label=above:$x_i$] {};
  \node[main] (y) [below=of x,label=below:$y_i$] { };
  \node[main, fill = black!10] (l) [right=of x,label=below:$\mathbf{L}_{ij}$] { };
  \node[main] (u) [above=of l,label=above:$u_j$] { };
  \node[main] (s) [left=of u,label=above:$\sigma$] { };
  \path (p) edge [connect] (W)
        (W) edge [connect] (y)
		(x) edge [connect] (y)
		(x) edge [connect] (l)
        (y) edge [connect] (l)
        (u) edge [connect] (l)
        (s) edge [connect] (u);
  \node[rectangle, inner sep=0mm, fit= (x) (y) (l),label=below right:m] {};
  \node[rectangle, inner sep=4.4mm,draw=black!100, fit= (x) (y) (l)] {};
  \node[rectangle, inner sep=0mm, fit= (u) (l),label=below right:n, xshift=4mm, yshift=-13mm] {};
  \node[rectangle, inner sep=8mm, draw=black!100, fit = (u) (l)] {};
\end{tikzpicture}
\caption{Graphical model of the DALC framework.}
\label{fig:dalc}
\vspace{-0.05in}
\end{figure}

\subsection{The Overall Framework}

The overall framework is depicted as a graphical model in Figure~\ref{fig:dalc}. It combines both the previously introduced DL model and the learning-from-crowd model  in a unified Bayesian framework. The joint probability of the framework 
is given by:
\begin{equation}
\small
\begin{aligned}
& p(y_i, \mathbf{u}_j, \mathbf{L}_{ij} | \mathbf{x}_i, K,  \sigma) \\
= & \underbrace{\int p(y_i|\mathbf{x}_i, \mathcal{W})p(\mathcal{W}|K)d\mathcal{W}}_\text{The Bayesian DL model} \underbrace{ \vphantom{\int} p(\mathbf{L}_{ij} | \mathbf{x}_i, \mathbf{u}_j, y_i) p(\mathbf{u_j}|\sigma)}_\text{The LFTC model}
\end{aligned}
\label{equ:dalc}
\end{equation}
\noindent where for the DL model we use the full Bayesian treatment for the parameter $\mathcal{W}$, which is critical for representing model uncertainty in active learning; while for LFTC it is sufficient to learn annotator expertise $\mathbf{u}_j$ by point estimates, therefore we use the prior distribution as a regularizer.

The DALC framework described above can learn the expertise of annotators across different data samples; simultaneously, it enables DL models to learn from sparse and noisy annotations. The learning process will be presented in the next section. Once DALC learns on an initial set of crowd annotations, it can actively select informative data samples and annotators with high-expertise for these samples to execute the annotation task. These annotated data samples will then be used to retrain the DL model, so as to improve model performance. With such an active learning procedure, the DL model is expected to reach the optimal performance with minimum amount annotation effort from the crowd. 

\smallskip
\noindent\textbf{Active Learning.} 
In active learning, DALC selects the $k$ most informative data samples and the annotator with the highest expertise for each of these data samples. The expertise of an annotator $u_j$ w.r.t. a data sample $\mathbf{x}_i$ is quantified by $\eta(\mathbf{x}_i, \mathbf{u}_j)$, as given by Equation~\ref{equ:eta}. The informativeness of a sample is defined by model uncertainty, quantified by Shannon entropy:
{\small
\begin{align}
\text{uncertainty}(x)  & = H[y|x, \mathcal{D}_{train}] \nonumber \\
& = - \sum_{c}p(y=c|x, \mathcal{D}_{train}) \log p(y=c|x, \mathcal{D}_{train}) \nonumber \\
& = - \sum_{c}(\frac{1}{T}\sum_t \hat{p}_c^t) \log (\frac{1}{T}\sum_t \hat{p}_c^t) 
\end{align}
}

\noindent where $\frac{1}{T}\sum_t \hat{p}_c^t$ is the averaged predicted probability of class $c$ for $x$, sampled $T$ times by Monte Carlo dropout. Note that $\mathcal{W}$ is marginalized in the above equation as in Equation~\ref{equ:pred_vi}.

\section{The Optimization Method}
\label{sec:opt}
This section describes the optimization method to learn the parameters of the DALC framework, including the DL parameters $\mathcal{W}$, and the LFTC parameters $\mathbf{u}_j$ ($1\leq j\leq n$) and $\mathbf{F}$. 

The parameters are learned by maximizing the likelihood of the observed annotations $\mathbf{L}$ given the data $\mathcal{X}$ and the annotators $\mathcal{U}$. Denoting all the parameters as $\Theta$, the optimization in the log space is formulated as:
%
\begin{equation}
\small
\begin{split}
\underset{\Theta}{\mathrm{argmax}} \log \prod_i \prod_j p(\mathbf{L}_{ij} | \mathbf{x}_i, K, \mathbf{u}_j) p(\mathbf{u}_j |\sigma) \\
= \underset{\Theta}{\mathrm{argmax}} \sum_i \sum_j \log \sum_{y_i} p(y_i, \mathbf{u}_j, \mathbf{L}_{ij}| \mathbf{x}_i, K, \sigma)
\end{split}
\label{equ:mle}
\end{equation}
where $p(y_i, \mathbf{u}_j, \mathbf{L}_{ij}| \mathbf{x}_i, K, \sigma)$ is given by Equation~\ref{equ:dalc}. The unknown variables $y_i$ ($1\leq i\leq m$) makes it computationally infeasible to directly solve the optimization problem. DALC employs the expectation maximization (EM) algorithm \cite{dempster1977maximum} to solve the problem.

 \begin{algorithm}[!t]\label{alg:1} 
	\caption{The EM Algorithm for DALC}
	\KwIn{data samples $\mathcal{X}$, annotation matrix $\mathbf{L}$, and $Iter_1$ }
	Initialize $p(y_i)\ (\forall 1\leq i\leq m), \mathbf{u}_j\ (\forall 1\leq j\leq n)$, $\mathcal{W}$, and $\mathbf{F}$\;
	\For{$t=1;t \le Iter_1;t++ $}  
	{
    	E-step: estimating $p(y_i)\ (\forall 1\leq i\leq m)$ by Equation~\ref{equ:e-step}\;
        M-step: updating $\mathcal{W}$ with back-propagation; \\
        \ \ \ \ \ \ \ \ \ \ \ \ \ updating $\mathbf{u}_j\ (\forall 1\leq j\leq n)$ and $\mathbf{F}$ by Algorithm~2\;
		\If{$\mathcal{L}$ has converged}
		{
			break\;
		}
	}
\end{algorithm}

\subsection{EM Algorithm for DALC}
The EM algorithm iteratively takes two steps, i.e., the E-step and the M-step. In each iteration, the E-step estimates the true labels given the current parameters; the M-step then updates the estimation of the parameters given the newly estimated true labels. 

\smallskip\noindent\textbf{E-step.} Using the Bayes rule, the true label is given by:
\begin{equation}
\small
\begin{split}
p(y_i) & \triangleq p(y_i|\mathbf{L}_{i:}, \mathbf{x}_i, K, \mathbf{u}_j) p(\mathbf{u}_j |\sigma) \\
& \propto \int p(y_i|\mathbf{x}_i, \mathcal{W}) p(\mathcal{W}|K)d\mathcal{W} \prod_j p(\mathbf{L}_{ij}|\mathbf{x}_i, \mathbf{u}_j, y_i) p(\mathbf{u}_j |\sigma)
\end{split}
\label{equ:e-step}
\end{equation}

Therefore, the true label is calculated based on both output of the Bayesian DL model and the LFTC model, as it is a function of the DL discriminator and the model of annotators. The prediction by the Bayesian DL model with the prior can be done as in Equation~\ref{equ:pred}, while the output of the LFTC model can be calculated by Equation~\ref{equ:bernoulli}.

\medskip\noindent\textbf{M-step.} Given the true label estimated by the E-step, we maximize the following likelihood function to estimate the parameters:
{\small
\begin{align}
\mathcal{L} = & \sum_i \sum_j \sum_{y_i} p(y_i) \log  p(y_i, \mathbf{u}_j, \mathbf{L}_{ij} | \mathbf{x}_i, K, \sigma) \nonumber \\
= &  \sum_i \sum_j \sum_{y_i} p(y_i)  \log  \bigg[ \int p(y_i|\mathbf{x}_i, \mathcal{W}) p(\mathcal{W}|K)d\mathcal{W}  \bigg. \nonumber \\
& \bigg. \times p(\mathbf{L}_{ij}|\mathbf{x}_i, \mathbf{u}_j, y_i) p(\mathbf{u}_j|\sigma) \bigg] \nonumber \\
 = & \sum_i \sum_j \sum_{y_i} p(y_i) \int \log p(y_i|\mathbf{x}_i, \mathcal{W})  p(\mathcal{W}|K) d\mathcal{W} \label{equ:m-step-dl}  \\
   & + \sum_i \sum_j \sum_{y_i} p(y_i) \left[\log p(\mathbf{L}_{ij}|y_i, \mathbf{x}_i, \mathbf{u}_j) + \log p(\mathbf{u}_j|\sigma)\right]  \label{equ:m-step-lftc} 
\end{align}
}

\noindent where $p(y)$ is obtained by E-step (Equation \ref{equ:e-step}). With the above equation, M-step can therefore be decomposed to two parts, namely, Equation~\ref{equ:m-step-dl} and \ref{equ:m-step-lftc}, which are independent from each other. The first part  (i.e., Equation~\ref{equ:m-step-dl}) is exactly the same as the objective function for training a Bayesian DL model, i.e., the cross-entropy loss function. It therefore can be optimized using the standard back-propagation method.  The second part (i.e., Equation~\ref{equ:m-step-lftc}) optimizes the LFTC model, which can be solved via a stochastic gradient ascent (SGA) method that will be given in the next subsection. The overall algorithm is given in Algorithm~1.

\begin{algorithm}[!t]\label{alg:2} 
	\caption{Learning the LFTC Model}
	\KwIn{data samples $\mathcal{X}$, current estimation of true labels $p(y_i)\ (\forall 1\leq i\leq m)$, annotation matrix $\mathbf{L}$, $d, \lambda, \gamma, Iter_2$ }
	Initialize $\mathbf{u}_j\ (\forall 1\leq j\leq n)$ and $\mathbf{F}$\;
	\For{$t=1;t \le Iter_2;t++ $}  
	{
		\ForEach{$\mathbf{L}_{ij} (1\leq i\leq m, 1\leq j\leq n)$}
        {
        	$\mathbf{u}_j \leftarrow \mathbf{u}_j  + \gamma \frac{\partial \mathcal{J}_{ij}}{\partial \mathbf{u}_j}$ by using Equation~\ref{equ:gra_u}\;
            $\mathbf{F} \leftarrow \mathbf{F} + \gamma 	\frac{\partial\mathcal{J}_{ij}}{\partial \mathbf{F}}$ by using Equation~\ref{equ:gra_F}\;
        }
		\If{$\mathcal{J}$ has converged}
		{
			break\;
		}
	}
\end{algorithm}

\subsection{Learning the LFTC Parameters}

Learning the LFTC parameters is equivalent to solving the following optimization problem:
{\small
\begin{align}
\underset{\mathbf{u}_j (1\leq j\leq n), \mathbf{F}}{\mathrm{max}} \mathcal{J} = & \sum_i \sum_j \underbrace{\sum_{y_i} p(y_i)  \left[\log p(\mathbf{L}_{ij}|y_i, \mathbf{x}_i, \mathbf{u}_j) +  \log p(\mathbf{u}_j|\sigma)\right]}_\text{$\mathcal{J}_{ij}$} \nonumber \\
  = & \sum_i \sum_j \underbrace{\sum_{y_i} p(y_i) \log p(\mathbf{L}_{ij}|y_i, \mathbf{x}_i, \mathbf{u}_j)}_\text{$\mathcal{J}_1$: log-likelihood}  \nonumber  \\
 & + \sum_i \sum_j \underbrace{\sum_{y_i} p(y_i) \log p(\mathbf{u}_j|\sigma)}_\text{$\mathcal{J}_2$: regularization}  
\end{align}
}

\noindent where we use $\mathcal{J}_{ij}$ to denote the objective function for learning parameters from a single annotation, and $\mathcal{J}_1, \mathcal{J}_2$ to respectively denote the log-likelihood and the regularization parts of $\mathcal{J}_{ij}$. Note that for the sake of clarity for deriving the gradients in the rest of this subsection, we formulate the optimization problem as a maximization problem instead of a minimization problem. 
The optimization problem is non-convex w.r.t. the parameters $\mathbf{u}_j$ ($1\leq j\leq n$) and $\mathbf{F}$. To solve the problem, we use alternative SGA, in which we alternately update $\mathbf{u}_j$ ($1\leq j\leq n$) and $\mathbf{F}$ until convergence.

To derive the gradient of the parameters for SGA, we first derive the gradient of $\eta(\mathbf{x}_i, \mathbf{u}_j)$ as below:
\begin{equation}
\small
\frac{\partial \mathcal{J}_{ij}}{\partial \eta(\mathbf{x}_i, \mathbf{u}_j)} = \frac{\partial \mathcal{J}_1}{\partial \eta(\mathbf{x}_i, \mathbf{u}_j)} = \frac{p(y_i=\mathbf{L}_{ij})}{\eta(\mathbf{x}_i, \mathbf{u}_j)} - \frac{p(y_i\neq\mathbf{L}_{ij})}{1-\eta(\mathbf{x}_i, \mathbf{u}_j)}
\end{equation}
The gradients of $\mathbf{u}_j$ and $\mathbf{F}$ w.r.t. $\eta(\mathbf{x}_i, \mathbf{u}_j)$ are given by:
\begin{equation}
\small
\begin{split}
\frac{\partial \eta(\mathbf{x}_i, \mathbf{u}_j)}{\partial \mathbf{u}_j}  & = \eta(\mathbf{x}_i, \mathbf{u}_j)(1-\eta(\mathbf{x}_i, \mathbf{u}_j)) \mathbf{F}\mathbf{x}_i \\
\frac{\partial \eta(\mathbf{x}_i, \mathbf{u}_j)}{\partial \mathbf{F}}  & = \eta(\mathbf{x}_i, \mathbf{u}_j)(1-\eta(\mathbf{x}_i, \mathbf{u}_j)) \mathbf{u}_j\mathbf{x}_i^\intercal
\end{split}
\end{equation}
In addition, the gradient of  $\mathbf{u}_j$ w.r.t. $\mathcal{J}_2$ is:
\begin{equation}
\small
\frac{\partial \mathcal{J}_2}{\partial \mathbf{u}_j} = - 2\lambda \mathbf{u}_j 
\end{equation}
where $\lambda = \frac{1}{\sigma^2}$. Finally, notice that both parts of the objective function  $\mathcal{J}_1$ and $\mathcal{J}_2$ are relevant to $\mathbf{u}_j$ ($1\leq j\leq n$), while only $\mathcal{J}_1$ is relevant to $\mathbf{F}$. Therefore, we have the following gradients for the LFTC parameters:
{\small
\begin{align}
\frac{\partial \mathcal{J}_{ij}}{\partial \mathbf{u}_j}  & = \frac{\partial \mathcal{J}_1}{\partial \eta(\mathbf{x}_i, \mathbf{u}_j)} \times \frac{\partial \eta(\mathbf{x}_i, \mathbf{u}_j)}{\partial \mathbf{u}_j} + \frac{\partial \mathcal{J}_2}{\partial \mathbf{u}_j} \label{equ:gra_u}\\
\frac{\partial\mathcal{J}_{ij}}{\partial \mathbf{F}} & = \frac{\partial \mathcal{J}_1}{\partial \eta(\mathbf{x}_i, \mathbf{u}_j)} \times \frac{\partial \eta(\mathbf{x}_i, \mathbf{u}_j)}{\partial \mathbf{F}} \label{equ:gra_F}
\end{align}
}

\noindent With these gradients, we can learn the LFTC parameters with alternative SGA. The overall algorithm is given in Algorithm~2.


\textfloatsep=5pt
\section{Experiments and Results}
\label{sec:experiment}
In this section, we conduct experiments to evaluate the performance of our proposed DALC framework. We aim at answering the following questions: 1) how effectively DALC can infer the true labels, and learn annotator expertise and annotation reliability, 2) how effectively DALC can train DL models from noisy and sparse crowd annotations, and 3) how effective DALC is in reducing the amount of annotations while training a high-performance DL model. 


\subsection{Experimental Settings}
\noindent\textbf{Datasets.} 
We use a real-world dataset from Amazon Alexa, which contains users' queries and their confirmation on  their query intents. The dataset contains 32,220 queries annotated by 10,006 users, who contribute a total of 49,958 annotations. The annotation matrix has a sparsity of 99.98\%. In addition to user annotated data, Alexa further contains a separate training dataset of over 50K queries with golden labels (i.e., labels with high agreement among users measured by conversion rates and judged by experts). 
%
%
To investigate the capability of DALC in inferring the true labels and uncovering the ground truth annotator expertise and annotation reliability, we created a synthetic dataset by simulating user annotations based on golden labels in the Alexa dataset. 



\smallskip
\noindent\textbf{Comparison Methods.} 
To demonstrate the performance of our proposed framework in model training, we compare the following multi-annotator methods: 
1) \textbf{MV}: infers the true labels with majority voting \cite{sheng2008get};
2) \textbf{LFC}: learns the true labels with learning-from-crowds \cite{yan2010modeling,yan2011active};
3) \textbf{STAL}: Self-taught learning-from-crowds proposed in \cite{fang2012self};
4) \textbf{DLC/LR}: our proposed framework where the target machine learning model is a logistic regression model;
5) \textbf{DLC/Sparse}: our proposed framework where the target model is a DL model, which learns annotator expertise without low-rank approximation -- the same method used by \cite{sheng2008get} and \cite{fang2012self};
6) \textbf{DLC}: our framework where the target machine learning model is a DL model that learns annotator expertise with low-rank approximation.
All compared methods, including ours, are aimed at training machine learning models on crowd annotations. We do not compare to methods designed only for output aggregation \cite{dawid1979maximum,whitehill2009whose}. Note that all existing methods (MV, LFC, and STAL) train a logistic regression model, and that DLC/LR, DLC/Sparse and DLC are our proposed framework variants without the active learning process.

To demonstrate the effectiveness of DALC in active learning, we compare the following variants: 
1) \textbf{RD+DLC}: randomly selects data samples and crowds for annotation;
2) \textbf{AD+DLC}: actively selects data samples while randomly selecting the crowds for annotation;
3) \textbf{AC+DLC}: randomly selects data samples while actively selecting the crowds for annotation;
4) \textbf{DALC}:  actively selects both data samples and high-expertise annotators.

\smallskip
\noindent\textbf{DL Model \& Parameter Settings.} We use a feed-forward neural network for the intent classification task in Alexa, which was proven to be more effective  in practice than recurrent networks~\cite{rumelhart1988learning}. 
Optimal parameters are empirically found based on a held-out validation set. We apply a grid search in \{0.0001, 0.001, 0.01, 0.1, 1\} for both the learning rate and regularization coefficient. A grid search in \{5, 10, 20, 50, 100\} is applied for both the number of components of Gaussian Mixture Model in STAL and the number of latent topics ($d$) in our framework. For the DL model, the dimension of the embedding is set to 200 and the number of hidden units is selected from the option set \{32, 64, 128, 256\}. The dropout value is validated from the option set \{0.10, 0.25, 0.50\}. Model training is performed using a RMSprop stochastic gradient descent optimization algorithm \cite{hinton2012rmsprop} with mini-batches of 128 data samples. 

\smallskip
\noindent\textbf{Evaluation Metrics.} We use accuracy and Area Under the ROC Curve (AUC) ~\cite{friedman2001elements} 
to measure the performance of the selected methods. 
Higher accuracy and AUC values indicate better performance.

\subsection{Learning from Targeted Crowds}
For the 50K queries with golden labels in Alexa, we simulate 50K users 
by creating for each of them a random vector $\mathbf{u}_j$ of size $d = 10$, whose elements are sampled from a uniform distribution $(-0.3, 0.6)$. We then generate a random matrix $\mathbf{F}$ whose elements are drawn from the same distribution. Based on $\mathbf{u}_j$ and $\mathbf{F}$ we calculate the annotation reliability for each query $\mathbf{x}_i$ as $\eta(\mathbf{x}_i, \mathbf{u_j}) = (1+\exp(\mathbf{u}_j^\intercal\mathbf{F}\mathbf{x}_i))^{-1}$. For individual users, the expertise is then represented by the percentage of correct annotations among all annotations they provide. 
Next, the annotations $\mathbf{L}_{ij}$ created by each user are assigned with either the same label as the golden label if $\eta(\mathbf{x}_i, \mathbf{u_j})>0.5$, or switched to a wrong label otherwise. Finally, we control for different annotation sparsity by randomly removing $1-\rho$ of the annotations for each user, where $\rho\in \{0.0001, 0.001, 0.01, 0.1\}$. 

\begin{figure}[!t]
        \centering
        \subfigure[Anno. Reliability Histogram]{
            \includegraphics[width=0.24\textwidth]{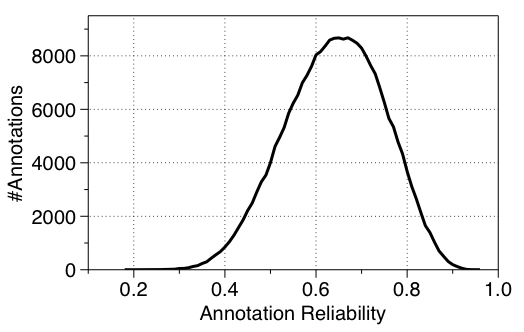}
            \label{}
        }%
        \subfigure[User Expertise Histogram]{
            \includegraphics[width=0.24\textwidth]{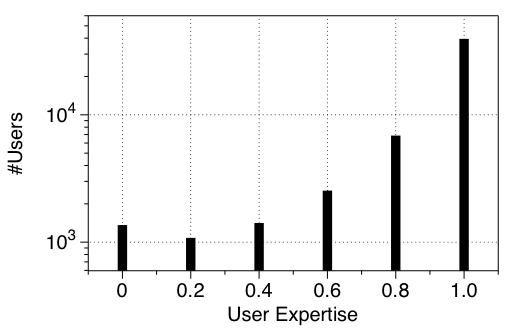}
            \label{}
        }\vspace{-0.05in}
        \subfigure[Anno. Reliability Comparison]{
            \includegraphics[width=0.24\textwidth]{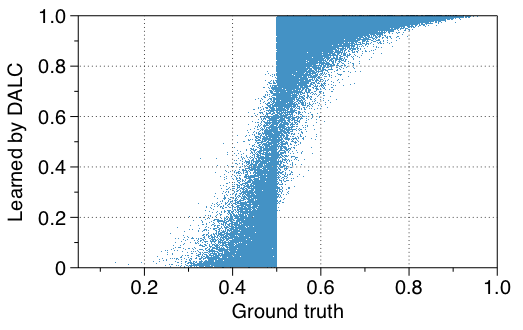}
            \label{}
        }%
        \subfigure[User Expertise Comparison]{
            \includegraphics[width=0.24\textwidth]{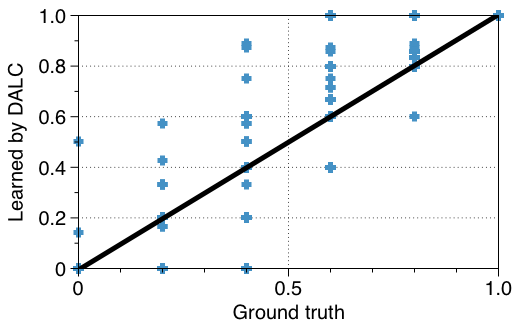}
            \label{}
        }
        \vspace{-0.05in}
        \caption{Upper figures are the histograms of  (a) annotation reliability and (b) user expertise in the synthetic dataset ($\rho=0.0001$); lower figures are the scatter plots of (c) annotation reliability learned by DALC vs. the ground truth, and (d) user expertise learned by DALC vs. the ground truth (fitted by a linear function).}
        \label{fig:simu_data}
\end{figure}

Figures~\ref{fig:simu_data} (a) and (b) show the histogram of annotation reliability and user expertise in the synthetic dataset with $\rho=0.0001$, respectively. It can be observed that annotation reliability is Gaussian-distributed, with a mean value being around 0.65. 
User expertise, in contrast, is not Gaussian. This is due to the \emph{discrepancy} between annotation reliability and annotation correctness: an annotation is correct as long as  $\eta(\mathbf{x}_i, \mathbf{u_j})$ is greater than $0.5$, either being $0.55$ or $0.95$. This will make an influence on the capability of DALC in learning annotation reliability, as we show later. 

\begin{table}[!t]
\centering
\addtolength{\tabcolsep}{-0.5mm}
\caption{Accuracy of labels inferred by DALC (row 2), correlation between the DALC learned annotation reliability, user expertise and the ground truth (rows 3 and 4).}\label{tab:simu_res}
\vspace{-0.1in}
\begin{tabular}{l|rrrrr}
\toprule
\textbf{Annotation Sparsity ($1-\rho$)}  & 0.0001 & 0.001 & 0.01 & 0.1 \\
\midrule
Inferred Label Accuracy  & 99.06\% & 99.96\% & 99.98\% & 100\% \\
\midrule
Anno. Reliability Correlation  & 0.5751 & 0.5891 & 0.6323 & 0.6768 \\
User Expertise Correlation & 0.9438 & 0.9921 & 0.9956 & 0.9998\\
\bottomrule
\end{tabular}
\end{table}

\smallskip
\noindent\textbf{Inferring True Labels.} Table~\ref{tab:simu_res} (row 2) reports the accuracy of the labels inferred by DALC w.r.t. the golden labels. DALC shows strong capability in inferring the true labels, even for highly sparse annotations (e.g., $\rho = 0.0001$). Recall that the annotations have an average reliability of $0.65$ (Figure~\ref{fig:simu_data} (a)). Despite this, we can observe that DALC can effectively recover the true labels from annotations that are both highly noisy and sparse.


\smallskip
\noindent\textbf{Annotation Reliability \& Annotator Expertise.}
Figures~\ref{fig:simu_data} (c) and (d) compare the user expertise and annotation reliability learned by DALC with the ground truth for the case when $\rho=0.0001$. For conciseness, we do not visualize datasets of other sparsity; however, similar observations as below can be obtained. 
First, DALC shows good performance in discriminating low reliability annotations from highly reliable ones; and, the reliability learned by DALC shows a positive correlation with the ground truth annotation reliability -- this is verified by Table~\ref{tab:simu_res} (row 3). However, we can observe that the correlation is not linear: the scatter plot of learned annotation reliability vs. the ground truth, as given in Figure~\ref{fig:simu_data} (c), exhibits a sigmoid shape. That is, the learned reliability for annotations whose ground truth reliability is greater than 0.5 can be as high as 1, though the lower bound increases for more reliable annotations. For annotations whose ground truth reliability is less than 0.5, the learned reliability can be as low as 0, thought the upper bound decreases for less reliable annotations. Such a phenomenon is due to the discrepancy between the annotation reliability and annotation correctness, as mentioned before. We note that the discrepancy, and as a result, the imperfectness in learning annotation reliability, is \emph{unavoidable} since DALC  learns the reliability of each single annotation by taking only annotations as the input, which is a challenging yet realistic task in real-applications. Finally, despite the imperfectness in learning reliability for individual annotation, DALC can accurately recover user expertise even for highly sparse annotations, as shown in  Table~\ref{tab:simu_res} (row 4) and  Figure~\ref{fig:simu_data} (d). 

In summary, DALC demonstrates strong capabilities in inferring the true labels and in learning annotator expertise, which are robust to data sparsity and noises.

\subsection{Performance in Model Training}
We now compare the performance of our proposed framework in model training with state-of-the-art multi-annotator model training methods. Table~\ref{tab:dl_res} compares our approach with benchmark methods in terms of accuracy and AUC on the Alexa dataset. 
From these results, we make the following observations.

Multi-annotator model training methods that use logistic regression to model the relationship between features and labels, including all existing methods (i.e., MV, LFC, and STAL) and DLC/LR, perform worse than  deep learning based model training methods. The large gap between the performance of these two types of methods -- e.g., DLC/Sparse outperforms DLC/LR by 6\% and 5\% in terms of accuracy and AUC, respectively -- clearly shows the advantage of deep learning models. 

Among logistic regression based methods, MV is outperformed by all the others, which implies the effectiveness of modeling annotator expertise in inferring true labels. LFC is outperformed by both STAL and DLC/LR. This is due to the fact that LFC suffers from the data sparsity problem in crowd annotations, as it takes the original feature representation of the data samples as input to model annotation reliability. In contrast, DLC/LR uncovers the low-rank structure of crowd annotations and maps original feature representations to low-dimensional representations for modeling annotation reliability, which greatly helps to resolve the annotation sparsity issue. Interestingly, STAL achieves comparable performance with DLC/LR. This is because while STAL is not specifically designed for coping with sparse annotations, the Gaussian Mixture Model employed by STAL reduces annotation dimensionality, which also helps to tackle the sparsity issue. 

For the two variants of our proposed framework that train a deep learning model, DLC significantly outperforms DLC/Sparse (Paired t-test, $p$-value $< 0.001$), which again implies the effectiveness of modeling the low-rank structure of crowd annotations to resolve the sparsity issue. Notably, the improvement brought by low-rank modeling of crowd annotations for deep learning models is larger than that for logistic regression. This on the one hand, can be attributed to the capability of deep learning models in capturing complex patterns in the data, which boosts the upper bound performance of deep learning models; on the other hand, the observation also suggests that sparse modeling is an effective way to improve the performance of deep learning models. 

\begin{table}[]
\centering
\caption{Performance of model trained using different multi-annotator methods on the Alexa dataset.}\label{tab:dl_res}
\vspace{-0.05in}
\begin{tabular}{c|cc}
\toprule
\textbf{Methods} & \textbf{Accuracy}& \textbf{AUC} \\
\midrule
MV & 0.6068 & 0.6302 \\
LFC & 0.6070  & 0.6304 \\
STAL & 0.6091 & 0.6325 \\
DLC/LR & 0.6093 & 0.6325 \\
DLC/Sparse & 0.6710 & 0.6860 \\
DLC & \textbf{0.6886} & \textbf{0.7235}  \\
\bottomrule
\end{tabular}
\end{table}

\begin{figure}[!t]
        \centering
        \subfigure{
            \includegraphics[width=0.23\textwidth]{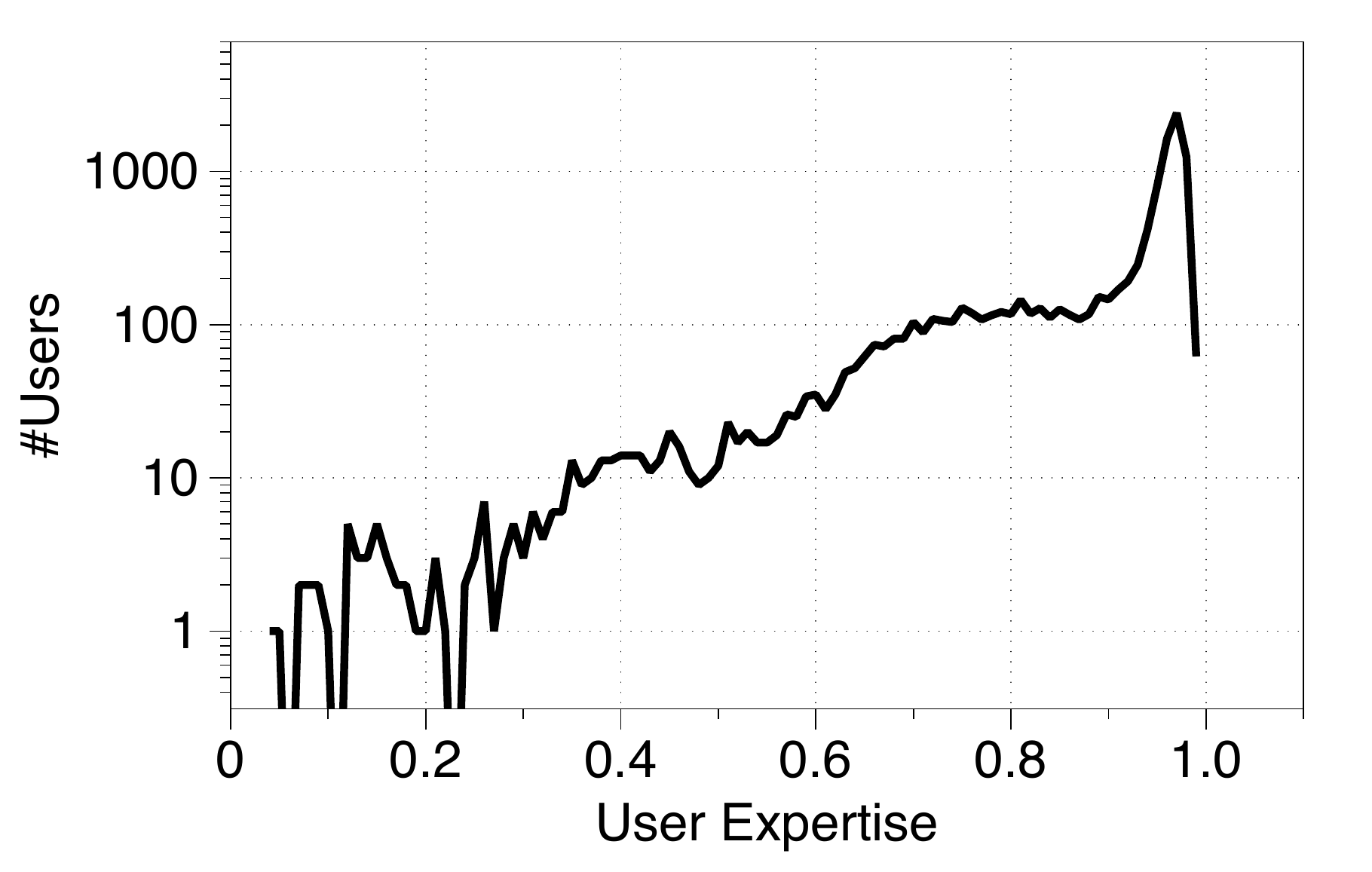}
        }\vspace{-0.1in}
        \caption{The learned expertise of users in the Alexa dataset.}
        \label{fig:dl_exp}
        \vspace{-0.1in}
\end{figure}
\begin{figure}[!t]
        \centering
        \subfigure[Impacts on accuracy]{
            \includegraphics[width=0.24\textwidth]{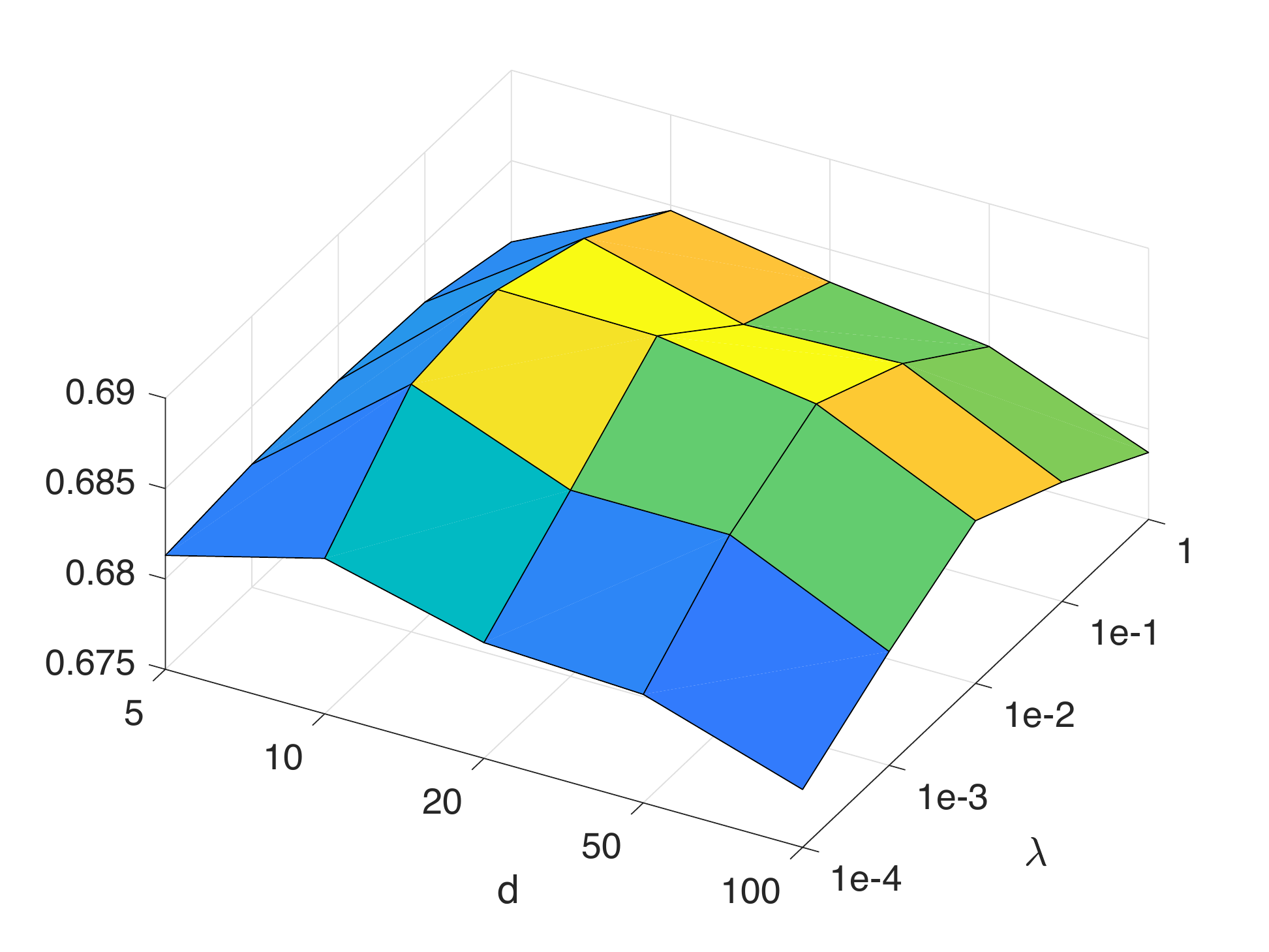}
            \label{}
        }%
		\subfigure[Impacts on AUC]{
            \includegraphics[width=0.24\textwidth]{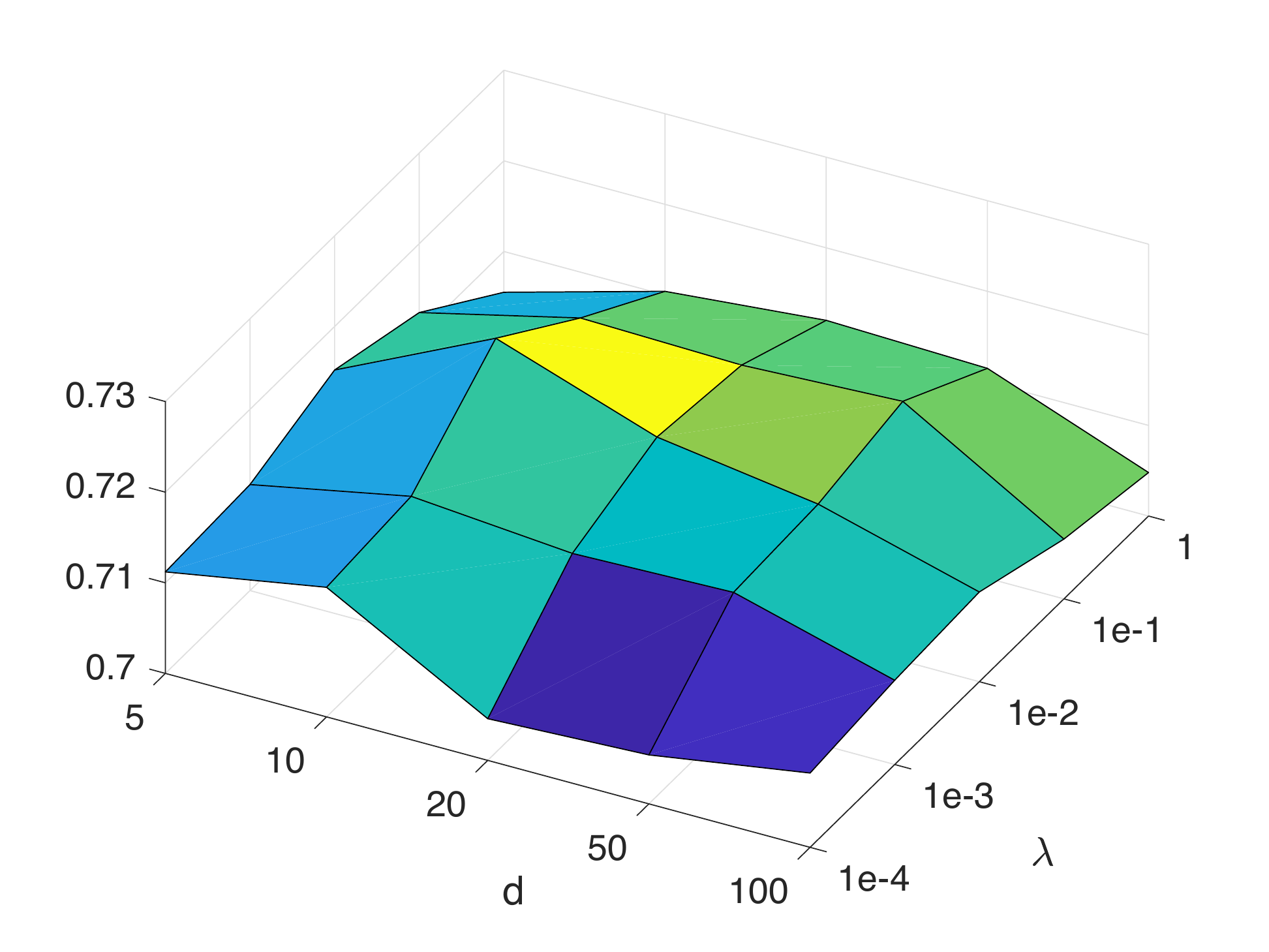}
            \label{}
        }
        \caption{The impacts of different hyper-parameter settings, including the dimensionality of the latent topics ($d$) and the regularization coefficient $\lambda$, on model training performance measured by accuracy (a) and AUC (b).}
        \label{fig:dl_pfm}
\end{figure}

\smallskip
\noindent\textbf{Distribution of Annotator Expertise.}  An advantage of our proposed framework is that it learns annotator expertise, which helps task owners understand the reliability of annotations. It can further be used for annotator selection in the active learning process, which will be discussed in the next subsection. The learned user expertise is shown in Figure~\ref{fig:dl_exp}. The average user expertise in Alexa is 0.88, indicating that users can generally provide high-quality annotations. 
Such annotation quality, however, is not fully applicable for Alexa in production, which motivates the need for modeling user expertise and annotation reliability. 


\smallskip
\noindent\textbf{Sensitivity to Hyper-parameters.} 
We investigate the impact of the number of latent topics $d$ and the regularization coefficient $\lambda$ which represents the regularization strength for annotator embeddings on the performance of the trained model. Results are shown in Figures~\ref{fig:dl_pfm} (a) and (b). As $d$ varies from small to large, the performance of the trained model first increases then decreases, with the maximum reached at $d=10$. Similar profile can be observed for $\lambda$, i.e., the performance of the trained model first increases then decrease when $\lambda$ varies from small to large, with the optimal setting reached at $\lambda=0.01$. The performance variations across different hyper-parameter settings suggest the need for parameter selection in the task; the similarity in performance variation across $d$ and $\lambda$ shows the robustness of DALC.

\subsection{Performance in Active Learning}
\begin{figure}[!t]
	\centering
	\subfigure[Performance by accuracy]{
            \includegraphics[width=0.23\textwidth]{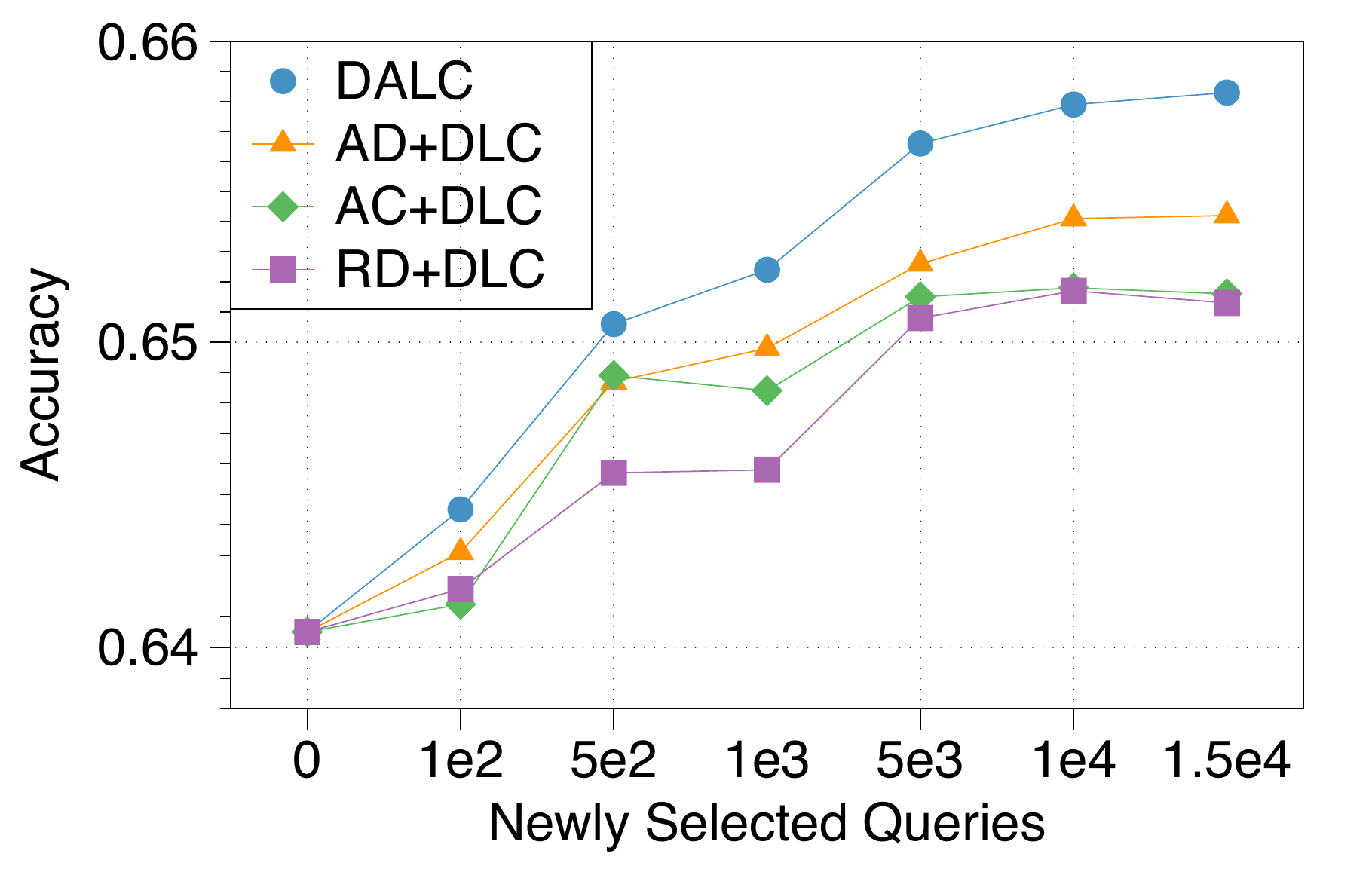}
            \label{}
        }%
        \subfigure[Performance by AUC]{
            \includegraphics[width=0.23\textwidth]{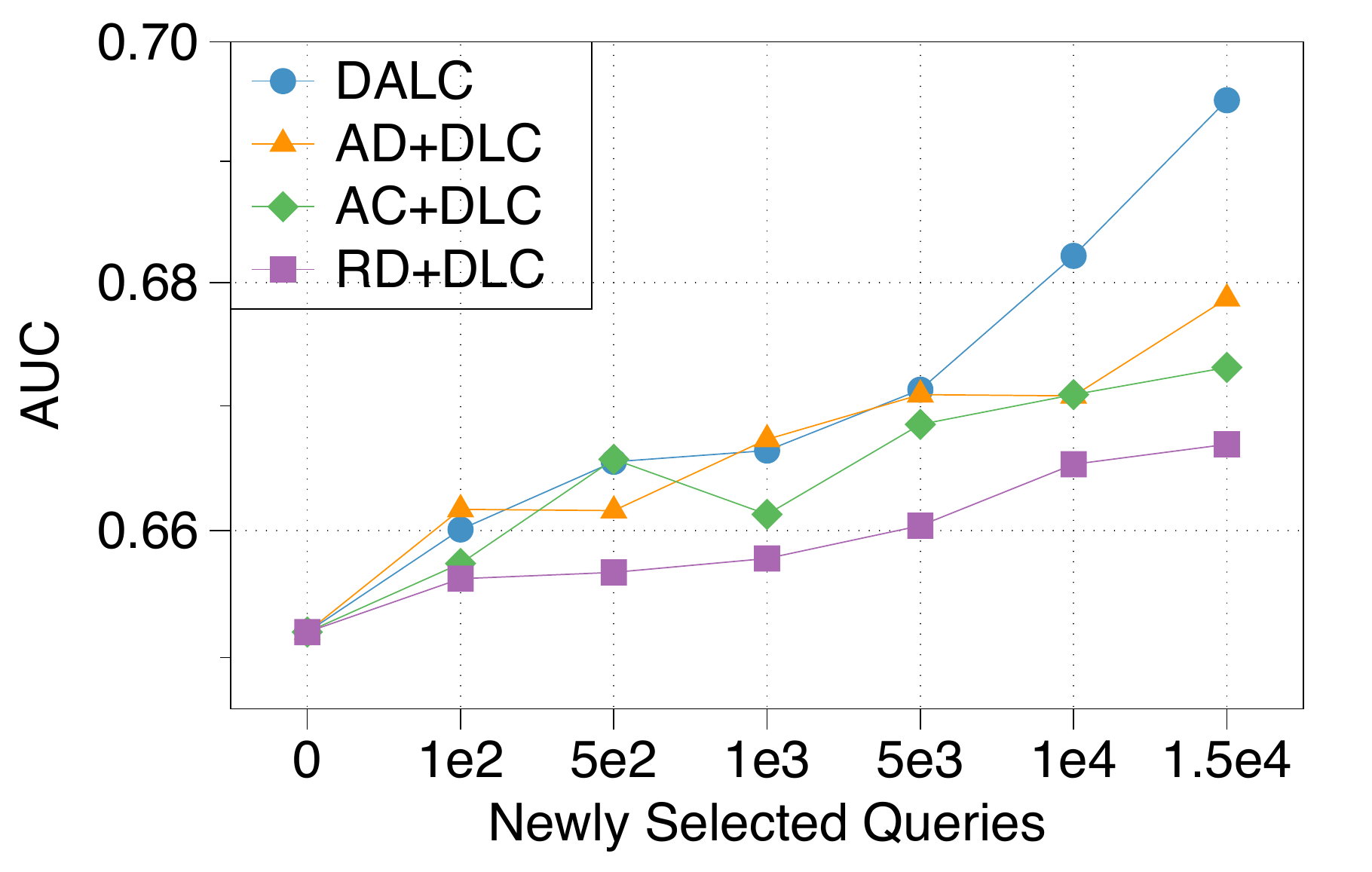}
            \label{}
         }
	\caption{Active learning performance of DALC variants on the Alexa dataset measured by accuracy (a) and AUC (b).} \label{fig:al_res}
\end{figure}

We now investigate the active learning performance of the compared methods on our real-world Alexa dataset. We separate the dataset into two parts, using 15K data samples for evaluating active learning performance and the rest of them to learn annotation reliability, which is used to guide the active selection of annotators. The active learning process is performed as follows. At each step, the model selects a certain amount of data samples with annotations, which is inserted to the existing training data to retrain the model; the performance of the retrained model is then measured using accuracy and AUC. 
Note that in each iteration the retrained model is used to select crowd annotations for the next iteration. Results are shown in Figure~\ref{fig:al_res}. 

With increased amount of newly selected data samples for model retraining, the performance generally increases. Both AC+DLC and AD+DLC outperform RD+DLC method, showing that the actively selected data samples and crowd annotations are both beneficial to model training. The fact that AD+DLC generally performs better than AC+DLC suggests that selecting the right data samples contributes more to the improvement of model performance. DALC outperforms all compared methods; when compared with the random selection method, it achieves an improvement of 1.03\% in Accuracy and 2.41\% in AUC when 15K data samples are selected. 
Interestingly, it can be observed that for DALC, a significant improvement margin over the compared methods is achieved when more than 10000 queries are selected. This confirms that deep learning models require large amount of data for model training, which further supports our motivation for using active learning to reduce crowd annotation efforts. 

Most importantly, we observe no significant difference of model performance between DALC and DLC trained on the same amount of data samples. In other words, by selecting annotations contributed by high-expertise users, we are able to reduce the number of annotations from 23.6K to 15K (reduced by 36.53\%) while preserving the same model performance. This clearly demonstrates the effectiveness of DALC in active learning. As a final remark, we note that our evaluation was limited to offline experiments; online settings require to further consider worker availability and possibly, the trade-off between labeling a new data sample and re-labeling an existing data sample \cite{lin2014re}, which is left for future work.

\vspace{0.07in}
\section{Conclusions}
We presented a Bayesian framework that unifies both the deep learning and the learning-from-targeted-crowd models, which are able to accurately learn annotator expertise and infer true labels from noisy and sparse crowd annotations. The framework enables any deep learning model to actively learn from targeted crowds, thus reducing the data annotation effort while reaching the optimal efficacy in training deep learning models. We extensively evaluated our framework in both synthetic and real-world datasets and showed that it consistently outperformed the state of the art. Our framework seamlessly connects deep learning models with targeted crowds; as a result, it opens up new research directions to explore worker/task modeling paradigm in crowdsourcing environments towards more advanced human-in-the-loop systems. 


\newpage
\widowpenalty
\clubpenalty
\bibliographystyle{ACM-Reference-Format}
\bibliography{sigproc}

\end{document}